\documentclass{article}
\usepackage{amsmath}
\usepackage{amsfonts}
\usepackage{graphicx}

\begin{document}

\title{\textbf{A mathematical framework of intelligence and consciousness based on Riemannian Geometry}}
\author{\Large Meng Lu\\\itshape{\small Peking University}\\ \small menglu@pku.edu.cn}

\date{}
\maketitle

\section*{Abstract}

Understanding intelligence is a central pursuit in neuroscience, cognitive science, and artificial intelligence. Intelligence encompasses learning, problem-solving, creativity, and even consciousness. Recent advancements in geometric analysis have revealed new insights into high-dimensional information representation and organisation, exposing intrinsic data structures and dynamic processes within neural and artificial systems. However, a comprehensive framework that unifies the static and dynamic aspects of intelligence is still lacking. This manuscript proposes a mathematical framework based on Riemannian geometry to describe the structure and dynamics of intelligence and consciousness. Intelligence elements are conceptualised as tokens embedded in a high-dimensional space. The learned token embeddings capture the interconnections of tokens across various scenarios and tasks, forming manifolds in the intelligence space. Thought flow is depicted as the sequential activation of tokens along geodesics within these manifolds. During the navigation of geodesics, consciousness, as a self-referential process, perceives the thought flow, evaluates it against predictions, and provides feedback through prediction errors, adjusting the geodesic: non-zero prediction errors, such as learning, lead to the restructuring of the curved manifolds, thus changing the geodesic of thought flow. This dynamic interaction integrates new information, evolves the geometry and facilitates learning. The geometry of intelligence guides consciousness, and consciousness structures the geometry of intelligence. By integrating geometric concepts, this proposed theory offers a unified, mathematically framework for describing the structure and dynamics of intelligence and consciousness. Applicable to biological and artificial intelligence, this framework may pave the way for future research and empirical validation.\\

\section*{Introduction}

Understanding intelligence, both in humans and artificial systems, has been a central pursuit in various fields including neuroscience, cognitive science, and artificial intelligence. Intelligence encompasses a wide range of cognitive abilities such as learning, problem-solving, creativity, and adaptation. These abilities are fundamental to how organisms interact with their environment, process information, and make decisions. In recent years, the geometric and topological tools have been developed and utilised to analyse the geometry of the high dimensional representation. These advancements have led to new insights into how information is processed and represented in neural and artificial systems (Hensel et al., 2021; Chung and Abbott 2021).\\

Researchers have used latent spaces to create low-dimensional representations of data manifolds, revealing their underlying geometrical structures (Reif et al., 2019; Marks and Tegmark 2023). Deep generative models like VAEs (Variational Autoencoders, Kingma and Welling, 2013) and GANs (Generative Adversarial Networks) have shown that these latent spaces can capture the curvature of learned manifolds, providing insights into intrinsic data structures (Arvanitidis et al., 2021; Chadebec and Allassonniere 2022). Additionally, algorithms for computing geodesic curves and parallel translation of tangent vectors allow for an intrinsic notion of distance and efficient navigation within these manifolds (Acosta et al., 2022). However, these models primarily describe the static data structure and lack mechanisms to account for the dynamics of how the latent space evolves over time. Another set of generative models, such as GPTs (Generative Pre-trained Transformers), utilise attention mechanisms to accumulate contextual information and autoregression to generate token sequences, which can be regarded as thought flow. This approach represents the dynamic aspect of intelligence. However, these models do not have an explicitly defined latent space in the same way that VAEs (Variational Autoencoders) and GANs (Generative Adversarial Networks) do. Consequently, the organisation and connections of knowledge or features within these models remain unclear.\\

In parallel with the advancement of geometric analysis in AI, the geometric representation of neural activities has been observed to efficiently encode behavioural variables and predict outcomes (DiCarlo and Cox 2007; Gao and Ganguli 2015; Vyas et al., 2020) by a variety of new tools and methods based on geometry (Jolliffe 1986; Tenenbaum et al., 2000; Roweis and Saul, 2000; Hinton and Roweis 2002; Mclnnes et al., 2018; Chaudhuri et al., 2019). For example, the hippocampus has been shown to encode both spatial and abstract variables within neural manifolds, serving as a common organising principle for storing declarative memory and generating cognitive maps (Dmitriy et al., 2017; Nieh et al., 2020). Apart from this, geometric representation has also been used in sensory recognition (Kobak et al., 2019; Okazawa et al., 2021; Stringer et al., 2019), motor control (Gallego et al., 2017) and decision making such as bayesian inference (Sohn et al., 2019). These studies highlight how geometric structures in the brain can facilitate decision-making and predictive coding through efficient integration of spatial and abstract information.\\

Building on the advancements in geometric analysis for both artificial and human intelligence, it is compelling to explore a general theory based on geometry (Lei et al., 2020; Bronstein et al., 2021; Ma et al., 2021) to describe the structure and dynamics of intelligence representation. This raises a few key questions: \textbf{1) If such a theory exists, what is the general form for representing information or features as elements within this framework, and what is their structure? 2) Under this representation and structure, what are the dynamics of the thought flow navigating within the structures formed by these representations? 3) Additionally, how does the structure of feature representation interact with the dynamic thought flow?} These questions highlight the need for a unified framework that encompasses both the static and dynamic aspects of intelligence, bridging the gap between information representation and cognitive processes.\\

A general theory of intelligence should encapsulate a range of properties emergent from intelligence, including learning, imagination, creative thinking, problem-solving, and the pinnacle of the intelligence hierarchy—consciousness. Consciousness, as an emergent property of complex cognitive processes (Seth and Bayne 2022), rests on the brain’s ability to sustain complex dynamics of constantly changing activity and connectivity between brain regions (Dehaene and Changeux, 2011; Hutchison et al., 2013; Barttfeld et al., 2015; Demertzi et al., 2019), indicating that consciousness influences and is influenced by the underlying structure of intelligence. Therefore, it is a necessary component of this general theory and can only be adequately explained and formulated within the comprehensive framework of intelligence. Understanding consciousness within this framework would provide a holistic view of how thought processes evolve and adapt, while explaining intelligence in the context of consciousness highlights the dynamic and self-referential nature of cognitive functions.\\

The author here proposes a theory of the geometry of intelligence that aims at addressing these questions via a comprehensive mathematical framework that describes the structure and dynamics of intelligence. This theory conceptualises elements of intelligence as tokens embedded in a high-dimensional space of intelligence, forming manifolds with geometric properties such as curvature. These manifolds capture both the static feature distributions and dynamic sequences of activation, reflecting the complexity and interconnections within cognitive processes. In the dynamics of intelligence and consciousness, sequential token activation forms thought flow, moving along manifold geodesics. The direction of this sequence, the tangent vector of this geodesic, is determined by the manifold's intrinsic geometry in terms of Riemannian geometry. Cognitively, the direction of thought flow is guided by contextual information from past tokens. Consciousness perceives and evaluates thought flow, providing feedback via prediction errors. When prediction error is zero, navigation follows the manifold's structure as a geodesic, representing free thought flow state with no consciousness perturbation. Typically, non-zero prediction errors and external input reflect the process of accepting new information and integrating it into the existing manifold, thereby evolving the intelligence space, representing process such as learning.\\

In summary, the geometry of intelligence guides how the consciousness navigate, the consciousness dictates how the geometry of intelligence evolves. By integrating geometric and topological concepts, this theory offers a novel and mathematically rigorous framework to describe the structure and dynamics of intelligence and consciousness, for both biological and machinery intelligence, paving the way for future research and empirical validation.\\

\section*{Background of Riemannian Geometry}

Riemannian geometry is a branch of differential geometry that studies smooth manifolds equipped with a Riemannian metric. This metric allows for the definition of various geometric concepts such as distances, angles, and curvatures on the manifold.

\subsection*{Manifold and Curvature}

A manifold \( \mathcal{M} \) is a topological space that locally resembles Euclidean space and is equipped with a smooth structure. The curvature of a Riemannian manifold is a measure of how much the manifold deviates from being flat. It is quantified using the Riemann curvature tensor \( R^\sigma_{\rho\mu\nu} \).

\subsection*{Metric Tensor}

The local geometry of a Riemannian manifold is defined by the metric tensor \( g_{\mu\nu} \). This tensor provides a way to measure distances and angles on the manifold. The components of the metric tensor in local coordinates are given by:
\[
g_{\mu\nu} = \frac{\partial x^\mu}{\partial u^\alpha} \frac{\partial x^\nu}{\partial u^\beta} g_{\alpha\beta}
\]
where \( x^\mu \) and \( x^\nu \) are local coordinates on the manifold, and \( g_{\alpha\beta} \) is the metric tensor in a different coordinate system.

\subsection*{Christoffel Symbols}

The Christoffel symbols \( \Gamma^\mu_{\nu\lambda} \) are derived from the metric tensor and represent the connection coefficients:
\[
\Gamma^\mu_{\nu\lambda} = \frac{1}{2} g^{\mu\rho} \left( \frac{\partial g_{\rho\nu}}{\partial x^\lambda} + \frac{\partial g_{\rho\lambda}}{\partial x^\nu} - \frac{\partial g_{\nu\lambda}}{\partial x^\rho} \right)
\]

\subsection*{Curvature Tensor}

The Riemann curvature tensor \( R^\rho_{\sigma\mu\nu} \) is a measure of the manifold’s curvature:
\[
R^\rho_{\sigma\mu\nu} = \partial_\mu \Gamma^\rho_{\nu\sigma} - \partial_\nu \Gamma^\rho_{\mu\sigma} + \Gamma^\rho_{\mu\lambda} \Gamma^\lambda_{\nu\sigma} - \Gamma^\rho_{\nu\lambda} \Gamma^\lambda_{\mu\sigma}
\]

\subsection*{Geodesic Equation}

The geodesic equation describes the evolution of a point moving along the manifold. The coordinates \( \gamma^\mu(t) \) describe the position of the point on the manifold as a function of the parameter \( t \), which could represent time or another affine parameter. The geodesic equation is given by:
\[
\frac{d^2 \gamma^\mu(t)}{dt^2} + \Gamma^\mu_{\nu\lambda} \frac{d\gamma^\nu(t)}{dt} \frac{d\gamma^\lambda(t)}{dt} = 0
\]
This equation ensures that the path \( \gamma(t) \) is locally the shortest path between points on the manifold, accounting for the curvature defined by the Christoffel symbols.

\section*{Geometry of Intelligence}

\section{Tokens, embedding and manifold}

Tokens, as discrete units, effectively represent various types of information, such as words in a sentence or pixels in an image, capturing complex data in a manageable form. High-dimensional spaces capture intricate relationships between data points, with each dimension representing a different attribute, enabling rich and detailed representations. These spaces facilitate learning and representing underlying data manifolds, crucial for understanding data structure and tasks like interpolation and extrapolation. High-dimensional spaces also separate data points that are close in lower dimensions, aiding in classification, clustering, and retrieval.\\

Advances in multi-modal models like Visual-Language Models (VLMs) such as CLIP (Contrastive Language–Image Pre-training) (Radford et al., 2021) integrate visual and textual information into a unified embedding space, showcasing the feasibility of a common space for embedding diverse data types. This integration highlights the power of high-dimensional token spaces in capturing complex, multi-modal relationships, enhancing understanding and generation tasks across various domains, and advancing both artificial and human-like intelligence.\\

Tokens are transformed into high-dimensional embeddings, which reside on manifolds capturing the data’s underlying structure. The curvature of these manifolds reveals important characteristics about the data distribution, influencing model performance. This geometric perspective provides deep insights into data relationships and guides the development of robust and efficient machine learning models.\\

Mathematically, let a token \( t_i \) be represented as a point in a high-dimensional space:
\[
t_i \in \mathbb{R}^d
\]
where \( d \) is the dimensionality of the space.

The embedding of token \( t_i \) is given by:
\[
v_i = \phi(t_i)
\]
where \( \phi \) is a function mapping the token to its embedding in the high-dimensional space.

The collection of all embeddings forms a manifold \( \mathcal{M} \) within this high-dimensional space. The dimension of the manifold, denoted as \( \text{dim}(\mathcal{M}) \), is typically much lower than \( d \), capturing the intrinsic structure of the data:
\[
\text{dim}(\mathcal{M}) \ll d
\]

\section{Thought flow and Geodesic}

\textbf{Concept}: In cognitive science, we propose that without perturbation or external stimulation, the thought flow navigates naturally on curved manifolds, following the geometry of the space. This unperturbed path satisfies the definition of a geodesic. In this context, the geodesic path is represented as a sequence of tokens activated along this path, reflecting the natural trajectory of thought flow in the high-dimensional token space. \textbf{The perturbed thought flow that is "forced" by stimuli or external input to deviate from its geodesic, and is the most common scenario, will be analysed in the later section of the Consciousness.} The geodesic is the path that minimises the distance between points, analogous to the shortest path on a curved surface. In this theory, the geodesic $\gamma(t)$ represents the state of intelligence at time $t$.\\

\textbf{Mathematical Representation} in cognitive aspect:
\[
\gamma(t) = \{v'_1, v'_2, \ldots, v'_n\}\tag{1}
\]
$\gamma(t)$ here is represented as a sequence of sampled embeddings $\{v'_1, v'_2, \ldots, v'_n\}$.
 \( v'_i \) is a sampled embedding from the distribution of the corresponding token \( t_i \):
\[
v'_i \sim \mathcal{N}(v_i, \Sigma_i)\tag{2}
\]
  - \( v_i \) is the mean embedding.
  - \( \Sigma_i \) is the covariance matrix representing the variability around \( v_i \).\\

Sampling from the distribution \( \mathcal{N}(v_i, \Sigma_i) \) rather than using a specific vector \( v_i \) allows the model to capture the inherent variability and uncertainty associated with each token. This approach ensures that the model generalises better to new, unseen data by reflecting the natural noise and variations present in real-world information. It also makes the geodesic path more robust to outliers, providing a more stable and reliable thought flow. Random sampling promotes exploration within the high-dimensional space, which can lead to discovering new and potentially better paths, similar to the use of randomness in transformers through techniques like top-k sampling and temperature scaling to introduce variability and flexibility in generating sequences.\\

By incorporating randomness through sampling, the model better mimics human cognitive processes, which are inherently probabilistic and uncertain. This stochastic approach allows the application of probabilistic methods for analysing and optimising the geodesic paths, leading to more accurate, robust, and adaptable representations of intelligence. Integrating these elements into the geometric framework of intelligence provides a comprehensive understanding of how complex cognitive processes are structured and evolve over time.

\section{Tangent Vector and State Transition Function}

\textbf{Concept}: The tangent vector at the moving front of the geodesic represents the direction and rate of change of the thought flow at that point. The differentiation of the geodesic function represents the state transition function, describing how the state of intelligence evolves over time.\\

\textbf{Mathematical Representation}:
\[
v(t) = \frac{d\gamma(t)}{dt}\tag{3}
\]
where $\gamma(t)$ is now a sequence of sampled embeddings $\{v'_1, v'_2, \ldots, v'_n\}$.\\

Here, the geodesic \(\gamma(t) = \{v_1', v_2', \ldots, v_n'\}\) represents a continuous and smooth path composed of connected points, or tokens, within a smooth manifold. This smooth structure ensures that functions defined on it, including \(\gamma(t)\), are continuous and differentiable.\\

The geodesic \(\gamma(t)\) models the trajectory of thought flow, implying gradual transitions between tokens without abrupt jumps. The tangent vector \(v(t) = \frac{d\gamma(t)}{dt}\) represents the derivative of this path with respect to time, providing a rigorous mathematical description of the rate of change of the thought flow. This differentiation is grounded in the manifold's smoothness, allowing for well-defined tangent vectors that capture the instantaneous direction and speed of movement along the geodesic. Thus, \(v(t)\) characterises the dynamic state of the system at any given time.\\

By treating the geodesic as a smooth path on a smooth manifold, we ensure mathematical rigor, applying differential geometry to model the continuous evolution of intelligence. This framework enables a precise description of token evolution over time, with \(v(t)\) encapsulating both the state and transition dynamics within the high-dimensional space of intelligence.\\

\section{Attention Mechanism}

\textbf{Concept}: The attention mechanism computes the relevance of each token in the sequence with respect to the current token, determining how contextual information influences the next token.\\

\textbf{Mathematical Representation}:
\[
\alpha_{ij} = A(v'_i, v'_j)\tag{4}
\]
where $\alpha_{ij}$ are the attention weights, and $A$ is a general attention function that measures the relevance of token $j$ to token $i$.\\

The curvature and structure of the manifolds are results of the training process, capturing the intrinsic links, organisation, and distribution of tokens. The attention mechanism measures the contextual significance and tokens' correlation, which is geometrically represented by the manifold's curvature and structure. This creates a fundamental link between the attention mechanism and the tangent vector. In mathematical terms, while the tangent vector provides a local derivative (instantaneous change), the attention mechanism offers a global perspective by integrating the influence of prior tokens. This integrated influence aligns with the geodesic's learned curvatures, as both are products of the underlying manifold's geometry.\\

\section{Context Vector and Contextual Embedding}

\textbf{Concept}: The context vector is a weighted sum of the value vectors, determined by the attention weights. It represents the aggregated contextual information for a given token.\\

\textbf{Mathematical Representation}:
\[
c_i = \sum_j \alpha_{ij} (W\omega v'_j)\tag{5}
\]
Here, $c_i$ is the context vector for token $t_i$, $\alpha_{ij}$ are the attention weights, and $W\omega$ is the learned value matrix.

\section{Predicted Token}
\textbf{Concept}: The predicted token concept is derived from the context vector. It represents the next token in the thought flow, influenced by the contextual information.

This concept can be expressed mathematically by two different equations:\\

\textbf{I. Contextual Representation}:
   \[
   g(t) = \sigma(W\phi c(t) + b\phi)\tag{6}
   \]
   where \(\sigma\) is the activation function, \(W\phi\) is a weight matrix, \(c(t)\) is the context vector at time \(t\), and \(b\phi\) is a bias term.\\

\textbf{II. Geometric Representation}

The geometric representation of the predicted token is given by:

\[ g(t) = \int_{t - \Delta t}^{t} v(t) \, dt + g(t - \Delta t) \]

where \( v(t) \) represents the incremental change over time, and \( g(t - \Delta t) \) is the previous state of the token.

In these equations, \( g(t) \) captures the essence of the predicted token by combining the immediate context with the temporal evolution of the thought flow. The integral component in the second equation signifies the accumulation of changes over time, reflecting the dynamic nature of thought processes.

\subsection*{Weight Matrix $W_{\phi}$}
The weight matrix $W_{\phi}$ and bias vector $b_{\phi}$ perform a linear transformation on the context vector $c(t)$. This transformation adjusts the integrated information in $c(t)$ to a new representation that can be used to predict the next token. It helps in mapping the high-dimensional context vector to the appropriate dimensional space of the next token. In human cognition, this can be likened to how the brain integrates various pieces of information and then transforms this integrated information into a specific thought or action. The weights and biases represent how different aspects of the accumulated knowledge are emphasized or de-emphasized in forming the next step in the thought process.

\section*{Non-linear Activation Function ($\sigma$)}

The activation function $\sigma$ (such as ReLU or tanh) introduces non-linearity into the model. Non-linearity is crucial for capturing complex relationships between tokens. Without non-linearity, the model would be limited to linear mappings, which cannot effectively represent the intricacies of thought processes. In human cognition, non-linear processing allows for the flexibility and complexity required in thinking, decision-making, and creativity. It allows the brain to process information in a way that is not simply additive or subtractive but can involve more complex interactions.\\

While the specific formulation provided is inspired by machine learning models like Transformers, the concepts of linear transformation and non-linear activation are fundamental to both human and machine intelligence.

\subsection*{Human Intelligence}
In the human brain, neurons process inputs through synaptic weights (analogous to $W_{\phi}$) and biases, and the non-linear activation is similar to the way neurons fire based on a threshold. This enables the brain to perform complex cognitive tasks by integrating and transforming information in sophisticated ways.

\subsection*{Machine Intelligence}
In artificial neural networks, linear transformations and non-linear activations enable the model to learn and represent complex patterns in data. These principles are essential for building models that can generalise from training data to make accurate predictions or decisions.\\

\section{Consciousness}

Consciousness emerges from intelligence through a series of complex cognitive processes, including internal monitoring, self-reflection, and adaptive behaviour. The interplay between consciousness and intelligence is characterised by the enhancement of cognitive functions such as perception, learning, memory, and attention, facilitated by self-awareness. Advanced forms of intelligence enable the development of consciousness, which, in turn, augments the capabilities of intelligent systems.\\

Mathematically, consciousness can be conceptualised as a self-referential vector, representing a state that continuously references and updates itself based on both internal and external inputs. This self-referential nature involves the system perceiving itself, evaluating predictions, and providing feedback to itself, thereby influencing subsequent cognitive states or tokens. For an intelligent agent, there is no discrete external input; instead, all stimuli must be integrated into the consciousness to be perceived effectively. Consequently, the perception within consciousness is the integration of the internal thought flow and external input, enabling a dynamic and holistic influence on the cognitive processes that underpin intelligent behaviour. This integrative process ensures that consciousness remains a cohesive and adaptive system, capable of responding to complex and evolving environments.\\

To analyse consciousness further, we can break it down into three fundamental steps and model each step as a function of time \( t \), representing the system’s internal state. These steps constitute a complete cycle of consciousness, including perception, evaluation, and feedback. The transition of state from \( t \) to \( t + \Delta t \) is used to represent this entire cycle shown as below:\\

\begin{figure}[h]
    \centering
    \includegraphics[scale=0.3]{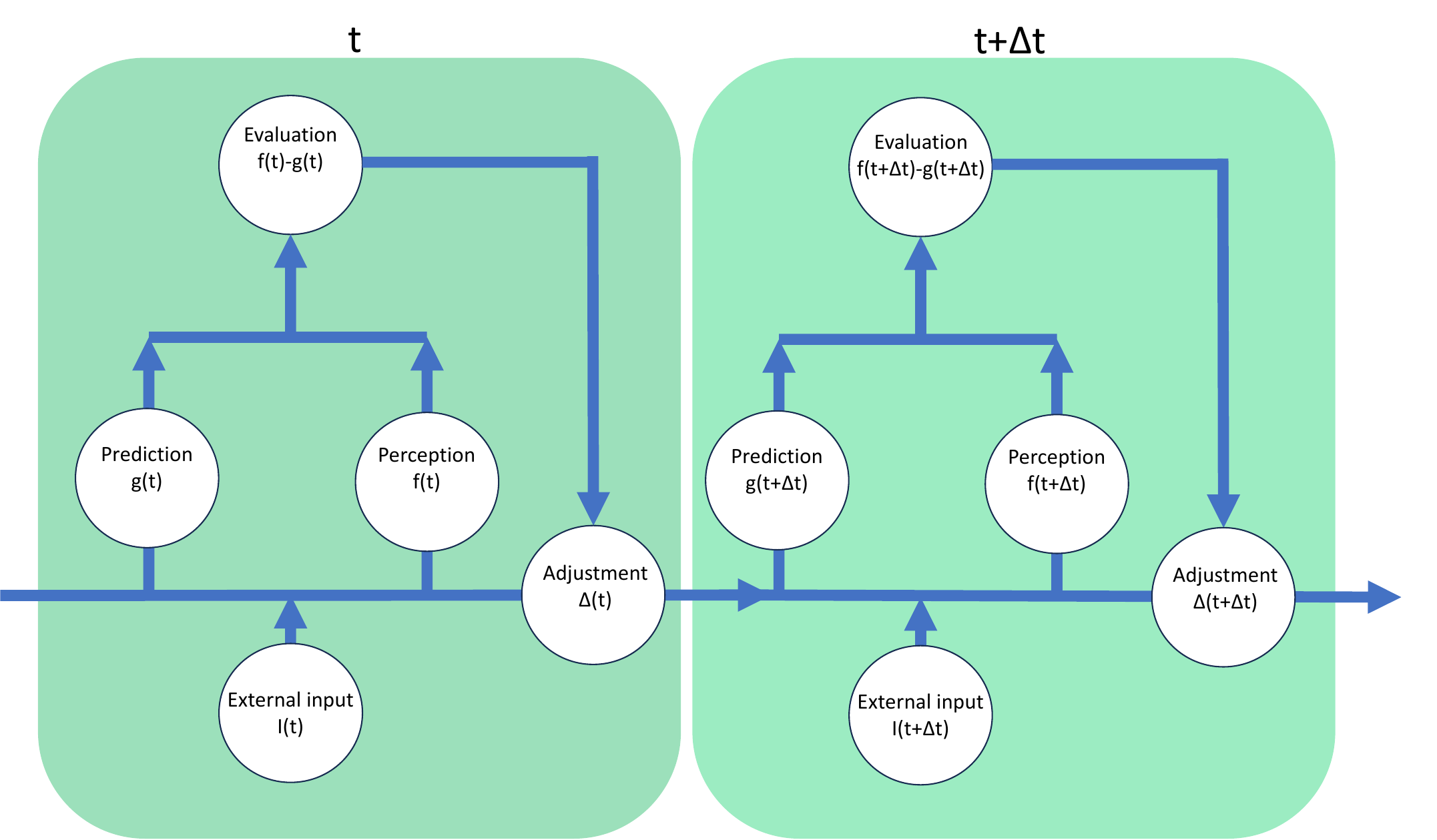}
    \caption{The flow chart illustrates the transition of consciousness cycle at t to the next cycle at \(t+\Delta t\): perception, prediction, evaluation, and adjustment. The left side represents the process at time \( t \), where external input \( I(t) \) is integrated with prediction \( g(t) \) to obtain the perception \( f(t) \). The evaluation \( f(t) - g(t) \) is used to adjust the thought flow. The right side represents the process at time \( t + \Delta t \), showing the continuous loop of feedback and adaptation.}
    \label{fig:consciousness_flow_chart}
\end{figure}

\begin{itemize}

\item \textbf{Perception}:The process by which sensory information is received and interpreted, modelled as a function of external stimuli and the current state. Perception integrates the internal thought flow and external input at time \( t \) to form the current state \( f(t) \).

\item \textbf{Evaluation}: The assessment of perceived information, including its emotional, cognitive, and contextual significance. This is modelled as the comparison between the current state \( f(t) \) and the predicted state \( g(t) \), resulting in the prediction error \( \Delta(t) = f(t) - g(t) \).

 \item \textbf{Feedback}: The system's response to the evaluation, influencing future perceptions, evaluations, and the overall state of consciousness. The feedback function \( \psi(\Delta(t)) \) adjusts the trajectory of the thought flow based on the prediction error, ensuring continuous adaptation and learning.

\end{itemize}

This cycle of perception, evaluation, and feedback ensures that consciousness remains a cohesive and adaptive system, capable of responding to complex and evolving environments. The transition from \( t \) to \( t + \Delta t \) represents one complete cycle of consciousness update, allowing for dynamic adjustments and the integration of new information into the system.\\

\subsection{Perception (Front Token)}

\textbf{Concept}: The front token is the current focus of the thought flow, representing the token that is being actively processed or considered at a given time.

Mathematically, we can express the front token \( f(t) \) as a function that captures the integration of internal thought flow and external input through the process of perception. The front token that is simply the position on the geodesic at time $t$ can then be modelled as:

\[ 
f(t) = P(\gamma(t), I(t))\tag{8} 
\]

where \( \gamma(t) \) is the internal thought flow at time \( t \), and \( I(t) \) is the input at time \( t \).

\subsection{Evaluation}

\textbf{Concept}: 
The prediction error is the difference between the front token and the predicted token. It measures the accuracy of the prediction, considering the capacities of perception, evaluation, and feedback.\\

\textbf{Mathematical Representation}
\[
\Delta(t) =  f(t) - g(t)\tag{9}
\]

\subsection{Feedback Function}

\textbf{Concept}: The feedback function adjusts the trajectory of the thought flow based on the prediction error. It helps in correcting the thought flow to improve future predictions.\\

\textbf{Mathematical Representation}:
\[
\psi(\Delta(t))\tag{10}
\]

Here, \(\psi\) is the feedback function that takes the prediction error as input and outputs the adjustment needed for the thought flow.

\section{Geodesic Equation with Feedback}

Based on the above analysis, we now can argue that the thought flow in an intelligent system can be modelled using the geodesic equation, which describes the path that \textbf{1) is determined by the geometry of the curved manifolds formed by token embeddings, 2) minimises the distance (or energy) in the manifold formed by the token embeddings}. The geodesic equation with feedback integrates the impact of consciousness by incorporating a feedback mechanism that modulates the trajectory of the thought flow based on prediction errors.\\

Mathematically, the geodesic equation with feedback is expressed as:    
\[
\frac{d^2 \gamma^\mu(t)}{dt^2} + \Gamma^\mu_{\nu\lambda} \frac{d \gamma^\nu(t)}{dt} \frac{d \gamma^\lambda(t)}{dt} = \kappa \cdot \frac{d^2 \psi(\Delta^\mu(t))}{dt^2} \tag{11}
\]

Here:
\begin{itemize}
    \item \(\frac{d^2 \gamma^\mu(t)}{dt^2}\) is the second derivative of the geodesic (acceleration).
    \item \(\Gamma^\mu_{\nu\lambda}\) are the Christoffel symbols representing the connection coefficients.
    \item \(\kappa \cdot \frac{d^2 \psi(\Delta^\mu(t))}{dt^2}\) is the modulated second derivative of the feedback function with respect to time.
   \end{itemize}

\textbf{Explanation}:
\begin{itemize}
    \item \(\kappa\) is the consciousness intensity index applied as a modulator to the prediction error. It represents the intensity with which the agent processes the prediction error. A higher \(\kappa\) indicates a more intense response to the prediction error.
   \end{itemize}

\textbf{Unit Consistency Check-Left Side of the Equation}: The left side of the geodesic equation represents the acceleration along the geodesic, with \(\gamma^\mu(t)\) being the position vector on the manifold:
\begin{itemize}
    \item \(\frac{d^2 \gamma^\mu(t)}{dt^2}\) is the second-order time derivative (acceleration) of the position vector \(\gamma^\mu(t)\), having units of \([L/T^2]\) (length per time squared).
    \item \(\Gamma^\mu_{\nu\lambda}\) are the Christoffel symbols, which are dimensionless since they are derived from the metric tensor, representing the connection coefficients.
    \item \(\frac{d \gamma^\nu(t)}{dt}\) and \(\frac{d \gamma^\lambda(t)}{dt}\) are first-order time derivatives (velocities), each having units of \([L/T]\) (length per time).
\end{itemize}

Thus, the left side has units of \([L/T^2]\) (length per time squared).\\

\textbf{Unit Consistency Check-Right Side of the Equation}: The right side involves the feedback term modulated by the consciousness intensity index \(\kappa\):
\begin{itemize}
    \item \(\Delta^\mu(t)\) represents the prediction error, which should have units of \([L]\) (length), assuming it is measured as a positional deviation in the manifold space.
    \item \(\psi(\Delta^\mu(t))\) is a function of the prediction error. To maintain unit consistency, \(\psi(\Delta^\mu(t))\) should also have units of \([L]\) (length).
    \item \(\frac{d^2 \psi(\Delta^\mu(t))}{dt^2}\) is the second-order time derivative of \(\psi(\Delta^\mu(t))\), having units of \([L/T^2]\) (length per time squared).
    \item \(\kappa\) is a dimensionless modulating factor.
\end{itemize}

Thus, the right side has units of \([L/T^2]\) (length per time squared).\\

\textbf{Zero Prediction Error}: When the prediction error is zero, \(\Delta^\mu(t) = 0\), the feedback term \(\kappa \cdot \frac{d^2 \psi(\Delta^\mu(t))}{dt^2}\) vanishes. In this scenario, the geodesic equation reduces to:
  \[
  \frac{d^2 \gamma^\mu(t)}{dt^2} + \Gamma^\mu_{\nu\lambda} \frac{d \gamma^\nu(t)}{dt} \frac{d \gamma^\lambda(t)}{dt} = 0
  \]
  This describes the natural geodesic path of the thought flow without any correction from prediction error feedback. The system follows its expected trajectory.\\

\textbf{Non-zero Prediction Error}: When there is a non-zero prediction error, \(\Delta^\mu(t) \neq 0\), the feedback term \(\kappa \cdot \frac{d^2 \psi(\Delta^\mu(t))}{dt^2}\) becomes significant. In this scenario, the geodesic equation incorporates the correction due to the prediction error:
  \[
  \frac{d^2 \gamma^\mu(t)}{dt^2} + \Gamma^\mu_{\nu\lambda} \frac{d \gamma^\nu(t)}{dt} \frac{d \gamma^\lambda(t)}{dt} = \kappa \cdot \frac{d^2 \psi(\Delta^\mu(t))}{dt^2}
  \]
  This modifies the trajectory of the thought flow, adjusting it based on the feedback from the prediction error. The system adapts its path as a result of the "force" acting on it, reflecting a dynamic response to the discrepancies between predicted and actual states.\\

\section{Competitive Activation and Consciousness Threshold}

\subsection{Competitive Process}

\textbf{Concept}: Multiple thought flows, generated by sampling from token distributions, compete based on their attention-derived scores. The flow with the highest score becomes part of the conscious experience.

\textbf{Mathematical Representation}:
\[
\text{Score}(\text{Thought Flow}_j) = f(\{\gamma_{j}(t)\})\tag{12}
\]
where \( \gamma_{j}(t) \) is the contextual embedding of the \( j \)-th thought flow at time \( t \).

\subsection{Consciousness Threshold}

\textbf{Concept}: The sequence with the highest score surpasses the consciousness threshold and becomes part of the conscious flow.

\textbf{Mathematical Representation}:
\[
\text{Conscious Flow} = \text{Thought Flow}_j \quad \text{if} \quad \text{Score}(\text{Thought Flow}_j) > \theta \tag{13}
\]

\textbf{Cognitive Science Aspect}: This competitive activation process mimics how human intelligence prioritises various thought processes. This concept is supported by earlier work suggesting that conscious awareness results from a competitive process where different cognitive inputs vie for limited attentional resources (Baars, 1988; Dehaene et al., 2001). Experimental studies have evidenced neural mechanisms of selective attention, demonstrating how stimuli compete for neural representation (Treue and Trujillo, 1999; Ruff and Driver, 2021). In the human brain, multiple streams of thought and potential actions are continually evaluated, with the most relevant or urgent thoughts reaching conscious awareness. This selective attention mechanism ensures the brain focuses on critical information, optimising cognitive resources and decision-making processes.\\

\textbf{Mathematical Aspect}: From a mathematical perspective, introducing competitive activation and consciousness thresholds helps to refine the model's efficiency and accuracy. By evaluating and selecting the thought flow with the highest score, the model can effectively prioritise the most relevant information, reducing noise and improving the robustness of predictions. This mechanism also introduces a layer of non-linearity and complexity that enhances the model's ability to simulate advanced cognitive processes, leading to the construction of more sophisticated AI models.

\section*{Summary and discussion}

\subsection*{Derivation of the Geometry Theory of Intelligence}

By representing thought flow as a geodesic in a high-dimensional space and incorporating mechanisms for perception, prediction, feedback, and random activation via token distributions, this theory provides a robust framework for modeling the structure and dynamics of intelligence. The mathematical formulations capture the continuous evolution of thought and the impact of contextual information, feedback, and randomness on this process.\\

The logical chain of deriving the theory of the geometry of intelligence begins with conceptualising elements of intelligence as tokens embedded in a high-dimensional space (Bjerke et al., 2023), which is grounded on the neuronal essemble as token and manifold, as well as token embedding in NLP and LLM. Each token represents discrete units of information. These tokens form manifolds, capturing both the static feature distributions and the dynamic sequences of activation. The curvature of these manifolds represents the organisation of tokens and the complexity of cognitive processes. Within this manifold, geodesics represent the natural and unperturbed paths of thought flow. The sequence of token activations follows these geodesic paths. Non-zero prediction error and input act as forces that deviate the thought flow from its native geodesic path to a new path. The result is the evolution of the geometry of the intelligence space, with token repositioning, curvature restructuring, new connection formation, and other changes. This formulates the learning process, as discussed later.\\

\subsection*{Comparison with the Working Mechanisms of Generative Models}

Given this theoretical foundation, it is crucial to compare this general framework of intelligence with the current state-of-the-art (SOTA) generative models. The SOTA generative models, such as VAEs, GANs, and transformer-based LLMs, represent the most advanced state of machine intelligence, often mimicking or even surpassing certain aspects of human intelligence, such as GPT-4 (Achiam et al., 2023) in general and Alphafold3 (Abramson et al., 2024) in specific task. These models may capture the essence of intelligence, including its principles and mechanisms, and provide insights into how complex cognitive processes can be modelled (Yang et al., 2024). Additionally, analysing human intelligence or brain function as a generative model (Friston and Price 2001; Spens and Burgess 2024) allows us to draw parallels between biological and artificial systems. By understanding these parallels, we can better appreciate the underlying structure and dynamics of intelligence in both realms, thereby enhancing our theoretical framework.\\

Pre-transformer generative models (VAEs and GANs) use a static, low-dimensional latent space to represent data, with geodesics serving to understand intrinsic distances and enable smooth interpolations on the data manifold (Bronstein et al., 2021; Chadebec and Allassonniere 2022; Acosta et al., 2023). Transformer-based models focus on dynamic token sequences without an explicit latent space, utilising attention mechanisms to determine the importance of input parts, where geodesics represent optimal sequences of token activations for coherent output generation.\\

The tokens (Radford et al., 2021; Lyu et al., 2023) with geometric representation forms manifolds that capture both the static distribution of features and the dynamic sequences of activation. The curvature of these manifolds reflects the complexity and interconnections within the cognitive processes. The static aspect of the theory encompasses the organisation of tokens based on their features, similar to the learned manifolds in VAEs and GANs, providing a foundational structure that captures the intrinsic relationships between different elements of intelligence.\\

Beyond static organisation, the dynamic sequences of token activation are crucial. Geodesics in this context represent the efficient pathways through which cognitive processes unfold over time. These sequences are learned during training and reflect the optimal transitions between different states of intelligence.\\

Integrating insights from both static and dynamic representations, the current theory benefits from the understanding of static feature representations provided by VAEs and GANs, extending the concept of a manifold to include dynamic interactions between tokens. The dynamic perspective offered by transformer models is crucial for modelling the temporal aspect of intelligence.\\

\subsection*{Interplay between token embeddings, curvature and geodesic}

The recent study on Claude 3 Sonnet (Templeton et al., 2024) provides valuable insights that can be seamlessly integrated into the theory of the geometry of intelligence. In this context, tokens are the fundamental units in the intelligence space, forming a manifold whose curvature determines the geodesic paths of thought flow. To fully leverage the findings from Claude 3 Sonnet, it is essential to elucidate the relationship between features and tokens and demonstrate how features can be represented by tokens within this theoretical framework.\\

Tokens, in this theory, are discrete units of information, such as words or images, embedded in a high-dimensional space to form points $t_i$. These embeddings $v_i = \phi(t_i)$ capture complex relationships and intrinsic structures, forming a manifold $M$. Features, on the other hand, are higher-level abstractions or patterns emerging from the interaction of multiple tokens. A feature could represent a concept like "Golden Gate Bridge," recognized through specific patterns in the embeddings of related tokens.\\

To bridge the gap between features and tokens, we can consider features as emergent properties arising from the combined activations of multiple tokens. A feature $F$ can be represented as a function of multiple token embeddings:
\[
F = f(v_1, v_2, \ldots, v_n)
\]
This function $f$ might be a weighted sum, convolution, or another aggregation mechanism capturing the interaction between tokens to form the feature. Each token embedding $v_i$ contributes to the feature's representation, and the feature's location in the high-dimensional space can be viewed as a region influenced by these tokens.\\

The curvature of the manifold $M$ reflects how token embeddings are organised. Features represent regions of high curvature where specific patterns or concepts are densely represented. Manipulating features effectively changes the manifold's organization and curvature, altering the geodesic paths.\\

\begin{itemize}
    \item \textbf{Curvature}: The curvature $\Gamma^\mu_{\nu\lambda}$ of the manifold is influenced by the distribution and interaction of token embeddings. Changes in feature activation alter this curvature.
    \item \textbf{Geodesics}: Geodesic paths, representing natural thought flow, change in response to feature manipulation, altering the trajectory through the high-dimensional space.
\end{itemize}

In this theory, the tangent vector represents the direction and rate of change of thought flow. The Claude 3 Sonnet study demonstrates that manipulating features changes the model's state transitions, aligning with changes in the tangent vector.

\begin{itemize}
    \item \textbf{Tangent Vectors}: Changes in the model's responses when a feature is manipulated correspond to changes in the tangent vector of thought flow.
    \item \textbf{State Transition}: New responses indicate a transition to different parts of the manifold, driven by changes in curvature from feature manipulation.
\end{itemize}

Feedback mechanisms in this theory adjust the manifold's structure based on non-zero prediction errors and external inputs, facilitating learning and adaptation. In Claude 3 Sonnet, feature manipulation by human/external input acts as a force to adjust the model's internal representations.\\

\begin{itemize}
    \item \textbf{Feedback}: Manipulating features changes the model's outputs, indicating a feedback mechanism that adjusts token embeddings.
    \item \textbf{Learning}: This process can be viewed as a form of learning, evolving the geometry of the intelligence space based on new inputs and feedback.
\end{itemize}

Using the mathematical framework of this theory, we can express the changes observed in the Claude 3 Sonnet study as follows:

\begin{itemize}
    \item \textbf{Feature Representation}:
    \[
    F = \sum_{i} w_i v_i
    \]
    Where $w_i$ are the weights determining each token embedding's contribution to the feature $F$.

    \item \textbf{Curvature and Geodesics}:
    The curvature $\Gamma^\mu_{\nu\lambda}$ of the manifold is influenced by these weighted embeddings. Manipulating a feature $F$ changes the weights $w_i$, thus altering the curvature and the geodesic paths $\gamma(t)$.

    \item \textbf{Geodesic Equation}:
    \[
    \frac{d^2 \gamma^\mu(t)}{dt^2} + \Gamma^\mu_{\nu\lambda} \frac{d\gamma^\nu(t)}{dt} \frac{d\gamma^\lambda(t)}{dt} = \kappa \frac{d^2 \psi(\Delta^\mu(t))}{dt^2}  \tag{11}
    \]
    Changes in $w_i$ due to feature manipulation impact the Christoffel symbols $\Gamma^\mu_{\nu\lambda}$, thereby altering the geodesic paths $\gamma(t)$.
\end{itemize}

By representing features as aggregated representations of token embeddings, we can integrate the concept of features within the theory of the geometry of intelligence. Manipulating features in the Claude 3 Sonnet study alters the embeddings and curvature of the manifold, changing the geodesic paths and affecting the thought flow. This framework provides a comprehensive explanation of how modifying features in AI models impacts their behavior, reflecting the dynamic interplay between geometry and thought flow in both artificial and biological intelligence.\\

\subsection*{Explaining the Advanced Functions and Properties of Intelligence}

\textbf{The Geometry of "Understanding"}\\

Understanding is a crucial cognitive process that allows for the coherent assimilation of new information, enabling smooth cognitive functioning and learning. In the context of our geometric framework, understanding involves the integration and stabilisation of new information within the manifold of intelligence. By comprehending the geometry of "understanding" and "misunderstanding," we can enhance our training models, improve understanding, and address misunderstanding more effectively. In our framework based on Riemannian geometry, these concepts are elucidated through token integration, geodesic navigation, and curvature dynamics.\\

Tokens are discrete units of information embedded in a high-dimensional space. Understanding occurs when new tokens are coherently integrated with existing ones, leading to smooth and stable manifolds that ensure seamless incorporation of new information. Misunderstanding arises when new tokens are poorly integrated, resulting in sparse or disconnected embeddings that form isolated clusters lacking coherence, characterised by high curvature, singularities, or undifferentiated regions, shown as Fig.2. This leads to unstable or erratic geodesics, where the thought flow exhibits abrupt changes, indicating a lack of coherence.\\

\begin{figure}[h]
    \centering
    \includegraphics[scale=0.3]{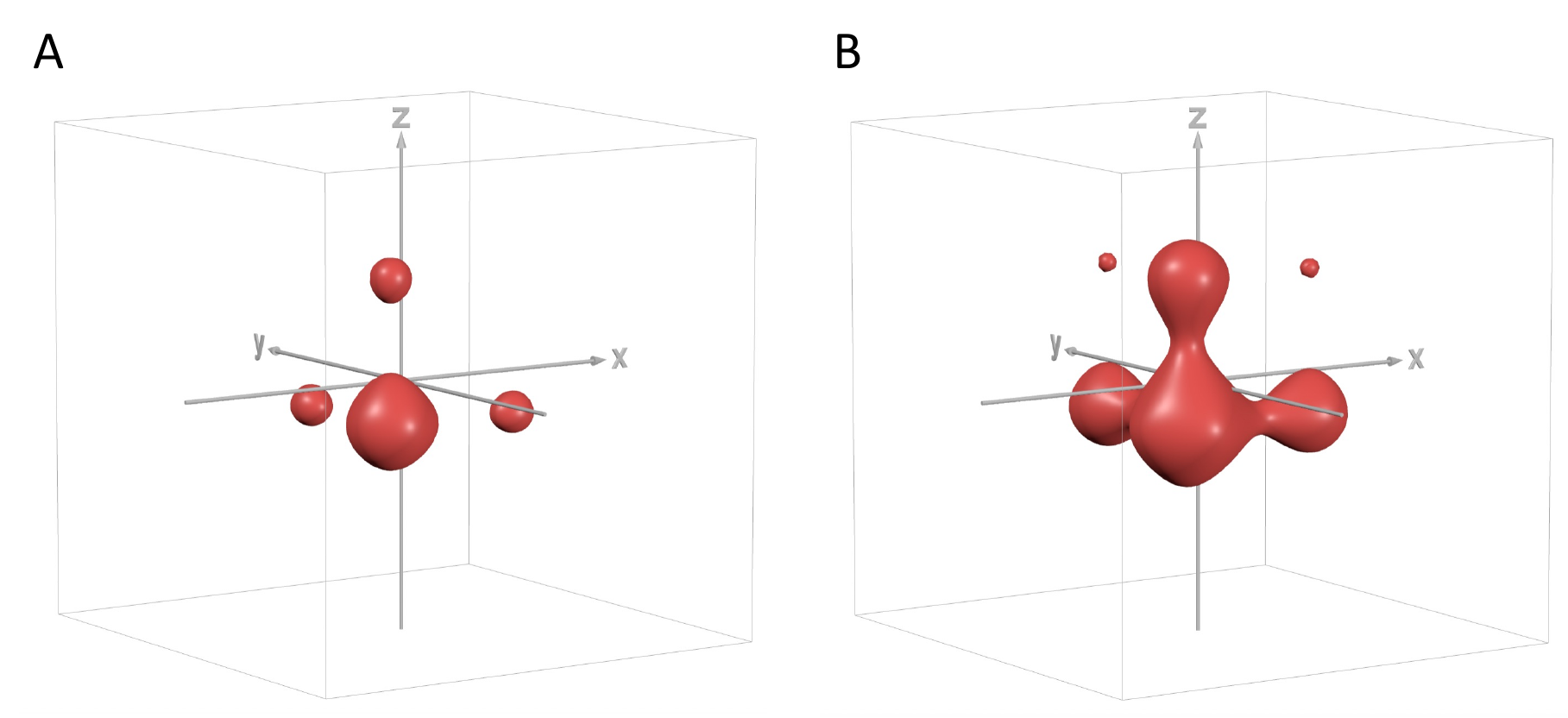}
    \caption{A simplified representation of manifolds in intelligence space. A. In the intelligence space, tokens are coherently connected to form smooth and stable manifolds. Some manifolds remain isolated, indicating that they are not explainable by others. No geodesic paths exist between these isolated manifolds. B. As learning progresses, the manifolds expand through the addition of new tokens. Consequently, more manifolds become interconnected, forming larger smooth and stable structures that facilitate longer and more complex geodesic navigation along them.}

    \label{fig:consciousness_flow_chart}
\end{figure}

\textbf{Curvature analysis} can identify regions likely to represent poor understanding or misunderstanding, allowing us to target and improve them specifically. In regions of misunderstanding, feedback loops fail to converge, leading to constant flux. In contrast, understanding achieves stable feedback loops, indicating well-integrated information. Erratic state transitions characterise misunderstanding, while smooth transitions indicate understanding. Visualising the manifold with techniques like t-SNE (Van der Maaten and Hinton 2008) or UMAP (Mclnnes et al., 2018) helps identify high-curvature, unstable regions indicating misunderstanding. Monitoring prediction errors also highlights areas needing further learning and integration.\\

After identifying regions of misunderstanding, we can specifically improve them using various methods. Active learning can seek additional information to integrate new tokens effectively. Strengthening feedback mechanisms can reduce prediction errors, stabilising the thought flow. Reinforcing connections between misunderstood and well-understood tokens leverages existing knowledge to facilitate better integration, converting misunderstanding into understanding.\\

By understanding and addressing misunderstanding in the geometric space, targeted strategies can improve the integration and coherence of new information, enhancing overall understanding. This approach ensures a robust and adaptable intelligence framework capable of evolving and improving over time.\\

\textbf{Imagination} is the ability to form new ideas, images, or concepts not present to the senses. Theories of imagination often emphasise its role in creative thinking and problem-solving, such as the notion of mental simulation, where the mind constructs possible scenarios to predict outcomes (Comrie et al., 2022; Hassabis and Maguire 2009; Schacter and Thakral 2024). In the framework of the geometry of intelligence, imagination can be understood as the activation of token sequences along geodesics that are not directly derived from immediate sensory input but rather from the manifold of stored experiences and abstract concepts. These imagined sequences $\gamma(t)$ can traverse novel paths in the high-dimensional space, allowing for the exploration of new ideas and creative solutions. The ability to navigate and synthesise these novel pathways showcases the flexibility and adaptability of the cognitive manifold in generating imaginative thoughts.\\

\textbf{Learning and experience} involve acquiring new knowledge or skills. Various theories explain learning as the strengthening of synaptic connections (Hebbian learning) (Hebb 1949; Sumner et al., 2020; Magee and Grienberger 2020) or the adaptation of cognitive structures (Piaget's theory of cognitive development) (Piaget 1953; Crone and Ridderinkhof 2011). In the geometry of intelligence, learning is conceptualised as the evolution of the manifold's structure through evaluation and feedback. The geodesic equation with feedback and input update:\\
\[
\frac{d^2 \gamma^\mu(t)}{dt^2} + \Gamma^\mu_{\nu\lambda} \frac{d \gamma^\nu(t)}{dt} \frac{d \gamma^\lambda(t)}{dt} = \kappa \cdot \frac{d^2 \psi(\Delta^\mu(t))}{dt^2} \tag{11}
\]

describes how the trajectory of thought flow adapts based on prediction errors. This adaptation, driven by the feedback function $\phi(\Delta(t))$ and input function $I(t)$, results in the modification of the token embeddings and the manifold's curvature. Thus, \textbf{geometry dictates how consciousness navigates, while consciousness guides the evolution of geometry}. This continuous interplay between the geometric structure and the dynamics of thought flow encapsulates the essence of learning, where the intelligence state evolves to reflect accumulated experiences and refined predictions. This relationship is exemplified in the study on Claude 3 Sonnet, where manipulating features leads to changes in the model's behavior.\\

When a feature's weight is amplified in Claude 3 Sonnet, it changes the model's geometric structure by altering the embeddings and their associations with other tokens. This modification affects the Christoffel symbols (\(\Gamma^\mu_{\nu\lambda}\)), which in turn alters the geodesic path of the thought flow. As a result, the model's responses (analogous to consciousness) change spontaneously, reflecting the new geometric configuration. This process aligns with the current theory where adjustments in geometry (feature weights and embeddings) directly influence the geodesic paths that consciousness follows, demonstrating the dynamic interplay between geometry and consciousness.\\

\textbf{Creative thinking} is the process of generating new, original ideas and solutions. It is often described as a recombination of existing knowledge in novel ways, facilitated by divergent thinking and cognitive flexibility (Beaty et al., 2016). In exisiting theories, creativity is linked to associative thinking and the ability to connect disparate concepts (Vartanian and Kaufman 2013; Suckling and Hoyer 2021). Within the geometry of intelligence, creative thinking can be modelled as the traversal of geodesics that connect diverse and previously unlinked regions of the cognitive manifold. The attention mechanism $\alpha_{ij} = A(v'_i, v'_j)$ plays a crucial role in dynamically weighting the relevance of various tokens, enabling the formation of unique and contextually rich combinations. By leveraging the manifold's structure and dynamically exploring new paths, the theory provides a robust explanation for the generation of innovative ideas and solutions, illustrating how the structural and dynamic aspects of intelligence facilitate creativity.\\

\textbf{Problem-solving} is the cognitive process of finding solutions to complex or challenging issues. Classical theories of problem-solving, such as those proposed by Newell and Simon (Newell and Simon 1972), focus on the stages of understanding the problem, generating potential solutions, and evaluating them. In the context of the geometry of intelligence, problem-solving involves navigating the manifold to identify and traverse geodesics that lead to effective solutions. The contextual embedding $c_{t} = \sum_j \alpha_{ij} (W_V v'_j)$ aggregates relevant information, guiding the thought flow toward potential solutions. The prediction and feedback loop $\Delta(t) = f(t) - g(t)$ ensures continuous refinement and adjustment of the cognitive trajectory, allowing for adaptive problem-solving. The integrated framework of geodesics, attention mechanisms, and feedback functions provides a comprehensive model that captures the dynamic and iterative nature of problem-solving, illustrating how the intelligence state transitions through various cognitive stages to arrive at a solution.\\

\textbf{In summary}, the mathematical framework of the geometry of intelligence provides a powerful tool for understanding key concepts, functions and properties of intelligence in both humans and machines. By modelling imagination, learning, creative thinking, and problem-solving as processes driven by the structured and dynamic activation of token sequences, this theory offers a unified explanation of cognitive phenomena, highlighting the intricate interplay between the geometry of intelligence and the flow of consciousness.\\

\section*{Nootations}
\subsection*{Tokens and Embeddings}

\begin{itemize}
    \item $t_i$: Token $i$
    \item $\phi$: Embedding function
    \item $v_i$: Embedded vector of token $t_i$
    \item $M$: Manifold formed by the collection of embeddings
\end{itemize}

\subsection*{Manifold and Curvature}

\begin{itemize}
    \item $g_{\mu\nu}$: Metric tensor defining the local geometry of the manifold
    \item $v_\mu, v_\nu$: Components of the embedding vectors
    \item $g_{\alpha\beta}$: Metric tensor in the original space
    \item $\Gamma^\mu_{\nu\lambda}$: Christoffel symbols representing connection coefficients
    \item $R^\rho_{\sigma\mu\nu}$: Riemann curvature tensor, a measure of the manifold’s curvature
\end{itemize}

\subsection*{Geodesic Equation with Feedback}

\begin{itemize}
    \item $\gamma(t)$: Geodesic path representing the state of intelligence at time $t$
    \item $f(t)$: Front token, current focus of the thought flow
    \item $I(t)$: Input at time $t$
    \item $\Delta(t)$: Prediction error
    \item $\psi$: Feedback function
    \item $\kappa$: Consciousness intensity index
    \item $\sigma$: Activation function
    \item $W_\omega$: Learned value matrix
    \item $W_\phi$: Weight matrix
    \item $b_\phi$: Bias term
\end{itemize}

\section*{Mathematical Equations}

\subsection*{Contextual Representation}
\begin{equation}
g(t) = \sigma(W_\phi c(t) + b_\phi)
\end{equation}

\subsection*{Geometric Representation}
\begin{equation}
g(t) = \int_{t - \Delta t}^{t} v(t) \, dt + g(t - \Delta t)
\end{equation}

\subsection*{Metric Tensor}
\begin{equation}
g_{\mu\nu} = \frac{\partial v_\mu}{\partial t_\alpha} \frac{\partial v_\nu}{\partial t_\beta} g_{\alpha\beta}
\end{equation}

\subsection*{Christoffel Symbols}
\begin{equation}
\Gamma^\mu_{\nu\lambda} = \frac{1}{2} g^{\mu\rho} \left( \frac{\partial g_{\rho\nu}}{\partial x^\lambda} + \frac{\partial g_{\rho\lambda}}{\partial x^\nu} - \frac{\partial g_{\nu\lambda}}{\partial x^\rho} \right)
\end{equation}

\subsection*{Riemann Curvature Tensor}
\begin{equation}
R^\rho_{\sigma\mu\nu} = \partial_\mu \Gamma^\rho_{\nu\sigma} - \partial_\nu \Gamma^\rho_{\mu\sigma} + \Gamma^\rho_{\mu\lambda} \Gamma^\lambda_{\nu\sigma} - \Gamma^\rho_{\nu\lambda} \Gamma^\lambda_{\mu\sigma}
\end{equation}

\subsection*{Geodesic Equation with Feedback}
\begin{equation}
\frac{d^2 \gamma^\mu(t)}{dt^2} + \Gamma^\mu_{\nu\lambda} \frac{d\gamma^\nu(t)}{dt} \frac{d\gamma^\lambda(t)}{dt} = \kappa \cdot \frac{d^2 \psi(\Delta^\mu(t))}{dt^2}
\end{equation}

\subsection*{Scenarios of Prediction Error}

\subsubsection*{Zero Prediction Error}
\begin{equation}
\Delta^\mu(t) = 0 \Rightarrow \frac{d^2 \gamma^\mu(t)}{dt^2} + \Gamma^\mu_{\nu\lambda} \frac{d\gamma^\nu(t)}{dt} \frac{d\gamma^\lambda(t)}{dt} = 0
\end{equation}

\subsubsection*{Non-zero Prediction Error}
\begin{equation}
\Delta^\mu(t) \neq 0 \Rightarrow \frac{d^2 \gamma^\mu(t)}{dt^2} + \Gamma^\mu_{\nu\lambda} \frac{d\gamma^\nu(t)}{dt} \frac{d\gamma^\lambda(t)}{dt} = \kappa \cdot \frac{d^2 \psi(\Delta^\mu(t))}{dt^2}
\end{equation}

  \end{document}